\documentclass[letterpaper, 10 pt, conference]{ieeeconf}  

\IEEEoverridecommandlockouts                              

\usepackage{graphics} 
\usepackage{graphicx}
\usepackage{epsfig} 
\usepackage{amsmath} 
\usepackage{amssymb}  
 \usepackage{algorithm}
\usepackage[compatible]{algpseudocode}
\algnewcommand\AAND{\textbf{ and }}
\algnewcommand\Or{\textbf{ or }}
\usepackage{color}
\usepackage{citesort}
\usepackage{flushend}
\usepackage{url}

\DeclareMathAlphabet{\pazocal}{OMS}{zplm}{m}{n}

\newcommand{\Xs}{\pazocal{X}}

\newcommand{\Rs}{\pazocal{R}}

\usepackage{array}
\newcolumntype{C}[1]{>{\centering\arraybackslash}p{#1}}
\newcolumntype{M}[1]{>{\raggedright\arraybackslash}p{#1}}

\usepackage{array} 
\newcolumntype{L}[1]{>{\raggedright\let\newline\\\arraybackslash\hspace{0pt}}m{#1}}	
\newcolumntype{S}[1]{>{\centering\let\newline\\\arraybackslash\hspace{0pt}}m{#1}}
\newcolumntype{R}[1]{>{\raggedleft\let\newline\\\arraybackslash\hspace{0pt}}m{#1}}

\newtheorem{problem}{Problem}

\makeatletter
\renewcommand*{\@opargbegintheorem}[3]{\trivlist
  \item[\hskip \labelsep{\bfseries #1\ #2}] \textbf{(#3)}\ }
\makeatother

\title{\LARGE \bf
L\'evy Flight Foraging Hypothesis-based Autonomous Memoryless Search Under Sparse Rewards}

\author{Christos Papachristos, and Kostas Alexis
\thanks{The authors are with the Autonomous Robots Lab, University of Nevada, Reno, 1664 N. Virginia, 89557, Reno, NV, USA
        {\tt\small kalexis@unr.edu}}%
}

\begin{document}

\maketitle
\thispagestyle{empty}
\pagestyle{empty}

\begin{abstract}
Autonomous robots are commonly tasked with the problem of area exploration and search for certain targets or artifacts of interest to be tracked. Traditionally, the problem formulation considered is that of complete search and thus - ideally - identification of all targets of interest. An important problem however which is not often addressed is that of time--efficient memoryless search under sparse rewards that may be worth visited any number of items. In this paper we specifically address the largely understudied problem of optimizing the ``time-of-arrival'' or ``time-of-detection'' to robotically search for sparsely distributed rewards (detect targets of interest) within large--scale environments and subject to memoryless exploration. At the core of the proposed solution is the fact that a search-based L\'evy walk consisting of a constant velocity search following a L\'evy flight path is optimal for searching sparse and randomly distributed target regions in the lack of map memory. A set of results accompany the presentation of the method, demonstrate its properties and justify the purpose of its use towards large--scale area exploration autonomy. 
\end{abstract}


\section{INTRODUCTION}

With an ever expanding application portfolio, robotic systems have established their role in a variety of critical domains. In response to this potential, the research community is pushing the limits with respect to the system capacity and overall intelligence. Most commonly, aerial robotics in particular are utilized as information gathering platforms exploiting their ability to seamlessly navigate without being subject to limitations of ground or surface locomotion~\cite{hunt2010acquisition,galceran2013survey,popovic2017multiresolution}. Despite the great progress of the research community, and although significant effort has been put in developing methods for autonomous exploration of complex and often unknown environments~\cite{SIP_AURO_2015,BABOOMS_ICRA_15,NBVP_ICRA_16,bircher2016receding,RHEM_ICRA_2017,SAEM_ICRA_2018,APST_MSC_2015,bourgault2002information,martinez2009bayesian,yamauchi1997frontier,zelinsky1992mobile,Thermal_ICUAS_2018,TUNNEL_AEROCONF_2018,papachristos2019autonomous,papachristos2016distributed,papachristos2016augmented,CD_ISVC_2016,mascarich2018radiation,dang2018autonomous,NIR_ICUAS_2017,thrun2004autonomous,stentz1994optimal}, little progress has been made for the specific consideration of rewards sparsity. 

However, a great variety of important applications relate to surveillance, monitoring and exploration missions within which information rewards may be particularly sparse. Indicative scenarios relate to a) search for survivors in a large-scale environment, and b) target detection and classification in the context of real-life Intelligence, Surveillance, Target Acquisition, and Reconnaissance missions. Furthermore, it is not unlikely that an autonomous robot deployed to execute such missions may be unable to maintain a full consistent pose estimate and the reconstruction of the map of its environment, a fact greatly emphasized in the framework of GPS-denied environments or applications. This lack of maintaining a consistent memory of the robot environment renders the problem of efficient information gathering significantly more difficult and sensitive. 

%
\begin{figure}[h!]
\centering
  \includegraphics[width=0.99\columnwidth]{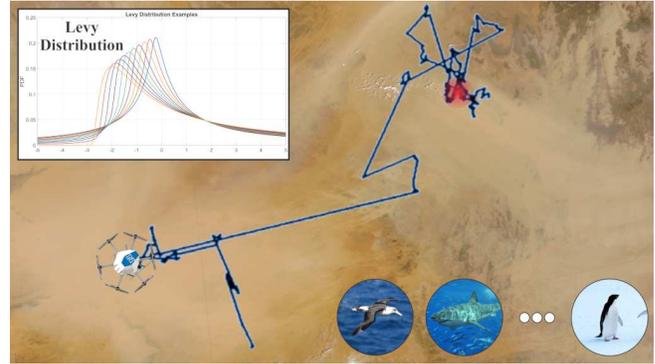}
\caption{The L\'evy Flight Foraging Hypothesis has been successful in modeling the exploration and food-search patterns observed in a variety of species. Motivated by this fact in this work we propose a memoryless large--scale area search algorithm subject to rewards sparsity. }
\label{fig:introfig}
\end{figure}
%

In attempt to provide a specific solution to the case of search under target/rewards sparsity and subject to inability to maintain a consistent memory of the environment, we looked to solutions prevailing in the animal kingdom. In a series of pioneering contributions~\cite{bartumeus2005animal,viswanathan2011physics,bartumeus2007levy,viswanathan1999optimizing,humphries2012foraging,benhamou2007many,reynolds2009levy,palyulin2014levy} it has been shown that a large set of species satisfy the so--called L\'evy Flight Foraging Hypothesis in their process of searching and identifying food regions or pray. The L\'evy Flight Foraging Hypothesis is based on the fact that a search-based L\'evy walk, consisting of a constant velocity search following a L\'evy flight path, is optimal for searching sparsely and randomly distributed target regions in the absence of map memory. Such L\'evy walk search patterns have been documented to model the movements of a set of species including - but not limited to - sharks, tuna, ocean sunfish, jellyfish, albatrosses, turtles, drosophila, and penguins~\cite{cole1995fractal,humphries2012foraging,gautestad2013levy}. This remarkable result, the ability of a simple inverse square power-law distribution to capture the behavior or at least certain animals when searching for pray, sets the basis for this work. With research in natural sciences continuously investigating and refining this model, in this work we look into a variation of this concept for the purposes of autonomous aerial robotic large-scale area search and target detection subject to rewards sparsity and lack of environment memory. A set of simulation studies are presented and indicate the potential of the proposed approach. Figure~\ref{fig:introfig} visualizes the concept of using L\'evy walks for search.

The remainder of this paper is structured as follows: The problem definition is given in Section~\ref{sec:ProbStat} and the proposed method is detailed in Section~\ref{sec:approach}. Evaluation studies are presented in Section~\ref{sec:evaluation}, followed by conclusions in Section~\ref{sec:concl}.

\section{PROBLEM FORMULATION}\label{sec:ProbStat}

When a robotic system, such as an autonomous unmanned air or ground vehicle, is tasked to enter a large area and search to detect targets of interest under a sparsity assumption, a traditional ``full coverage'', even if possible, procedure is not efficient when tasked to ensure fast arrival to the first detection case. In addition such solutions are admissible only when the ability for the robot to know its pose and map in a drift--free manner is possible, which is often not the case in a wide set of critical missions. However, this goal - to optimize the time of first detection - can be particularly important in the framework of a large set of critical applications in the civilian and military domains alike. Thus, a new set of methods has to be developed. We call the problem to be addressed as that of ``Memoryless Sequential Time-of-first-Arrival Optimized Search under Sparse Rewards''. In the framework of the problem considered, ``target'' or ``reward'' regions may be visited any number of times. 

\begin{problem}[Memoryless Sequential Time-of-first-Arrival Optimized Search under Sparse Rewards]\label{def:sparsesearch}
 Let $\Xs$ be the configuration space containing a set of sparse targets (rewards) $\Rs$ that are co-located on a small number of $N_C$ clusters $\mathbf{r}_i,~i=1,..,N_C$. Optimize the time-of-arrival to the first cluster $t_a$ and the associated rewards collection, from that to the next cluster and so on up to the $N_C$-th cluster. The robot is considered memoryless with respect to the map, while the target sites may be visited any number of times or the respective rewards may be depleted after the first ``hit''. 
\end{problem}

\section{PROPOSED APPROACH}\label{sec:approach}

To address this problem we look into the L\'evy Flight Foraging Hypothesis (LFFH)~\cite{viswanathan1999optimizing,bartumeus2005animal} and the extensive relevant literature in biological sciences~\cite{palyulin2014levy,reynolds2009levy,benhamou2007many,humphries2012foraging,bartumeus2007levy,viswanathan2011physics,bartumeus2005animal}. The pioneering work in~\cite{viswanathan1999optimizing} conducted studies analyzing experimental foraging data on selected insect, mammal and bird species and found that when the target sites are sparse and may be visited any number of times, an inverse square power-law distribution of flight lengths, corresponding to L\'evy flight motion, is an optimal strategy. As first detailed by~\cite{levandowsky1988swimming,schuster1996chemosensory}, a L\'evy distribution is advantageous at least when target sites are sparsely and randomly distributed, among others because the probability of returning to a previously visited site is smaller than, for example, a gaussian distribution. 

\subsection{L\'evy Flight Foraging Hypothesis}

\begin{figure}[h!]
\centering
  \includegraphics[width=0.99\columnwidth]{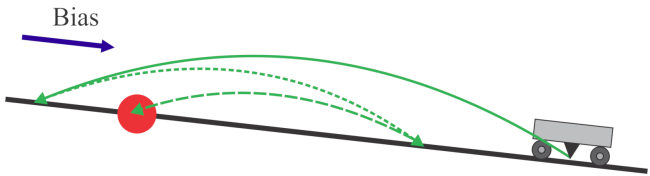}
\caption{Blind random search process in one dimension. A robotic agent is tasked to perform random jumps in its search space until it arrives to the target location to collect rewards. In case of a bias due to, for example, a drift away from the target, then the L\'evy Flight based random jumps may overshoot the target in a manner that the first arrival to the target is less efficient than the passage across the target. Such drifts can happen in real robot disturbances either due to external disturbances or even due to localization challenges. }
\label{fig:biasedsearch}
\end{figure}

The LFFH states that since L\'evy Flights and walks can optimize search efficiency, therefore natural selection should have led to adaptations for L\'evy flight-based foraging. For the probability distribution of a L\'evy flight it holds that:

\begin{eqnarray}\label{eq:levybasic}
p(x)\sim x^{-\mu}~\rightarrow~ \mathbf{P}(X>x)\sim x^{1-\mu}
\end{eqnarray}
where $1 < \mu \le 3$ and $x$ is the flight length. In this expression, the gaussian is the stable distribution for the special case where $\mu \ge 3$ as a result of the central limit theorem. On the other hand, for values $\mu \le 1$ it corresponds to probability distributions that can be normalized. Let us define the search efficiency function $\eta(\mu)$ to be the ratio of the number of target sites visited to the total distance traversed by the forager:

\begin{eqnarray}
\eta = \frac{1}{\left \langle x \right \rangle N}
\end{eqnarray}
where $N$ is the mean number of flights taken by a L\'evy forager while traveling between any two successive target regions. For the case of rewards sparsity, which in turn means that the average foraging length $\lambda$ is in general much larger than the ``detection radius'' (the area within which the agent can detect its target) $r_d$, then the work in~\cite{viswanathan1999optimizing} first showed that the optimal search policy is based on Equation~(\ref{eq:levybasic}) with

\begin{eqnarray}
\mu_{opt} = 2-\delta, \delta \approx \frac{1}{\ln(\lambda/r_d)} 
\end{eqnarray}
which in turn means that the optimal selection is to chose $\mu_{opt}=2$ when $\lambda/r_d$ is large but not known a priori or exactly. The associated variance of the agent jump lengths diverges, such that the resulting scale--free jump process features occasional, extremely long jumps. This is in contrast to what is observed in Brownian search with frequent returns to areas previously visited. As discussed in~\cite{shlesinger1986levy} these long jumps offered by such a L\'evy Flight model improve the efficiency of the search process by leading the randomly walking agent to statistically uncorrelated areas within its search area. For this model to refer to the optimal search under sparse rewards, three further assumptions are of importance, namely that a) the agent has no memory with respect to places previously visited, b) when rewards are within the detection range $r_d$ then they are sequentially collected before the agent forages to a new direction, and c) the jumps are ``unbiased'' by any possible drift. Notably, under such considerations the mean number of jumps $N_d$ taken to travel a mean distance $\lambda$ between two successive target regions scales on the basis of the following formula:

\begin{eqnarray}
N_d \approx (\lambda/r_d)^{\mu -1}
\end{eqnarray}
for any $1<\mu \le 3$ and $\mu$ representing the fractal dimension of the set of target regions. With respect to the importance of non-biased walks, a counter example is visualized in Figure~\ref{fig:biasedsearch} on the basis of the discussion in~\cite{palyulin2014levy}. 

%
%

\subsection{LFFH-based Search under Rewards Sparsity}

In this work, we propose an idealized model of target/rewards search in large areas subject to rewards sparsity, while assuming no memory with respect to places previously visited. Dynamics of predator-prey are ignored which is applicable either because the reward regions are static or because there is no knowledge about the underlying dynamics, or simply there is no clear way to incorporate such knowledge in the framework of a certain application. The proposed approach for LFFH-based search under rewards sparsity relies on the following switching behavior: 

\noindent \textbf{Mode A - LFFH Foraging:} If there is no target within a detection range $r_d$ then the foraging agent selects a random direction of distance $\ell_j$ sampled from the probability distribution in Equation~(\ref{eq:levybasic}). Once selected, the agent transits to the new point, while simultaneously constantly sampling to detect targets within $r_d$ along its way. If the agent does not detect any target, then it proceeds until it arrives to the point in distance $\ell_j$ and then chooses a new direction and a new length $\ell_{j+1}$ also from Equation~(\ref{eq:levybasic}). If the agent detects a target of interest then it switches to the next mode. During this first mode, the behavior of the forager is analogous to that of random walks whose mean-square displacement is proportional to the number of steps in any dimension of the agent configuration space. 

\noindent \textbf{Mode B - Local Reward Collection:} If a target lies within the agent's direct detection range $r_d$ then the system switches to a local behavior within which it randomly selects any of the available reward regions, visits it to collect it and continuous this process until there is no other immediate reward to collect. When this condition is met, the agent switches to the previous mode. 

\begin{figure}[h!]
\centering
  \includegraphics[width=0.99\columnwidth]{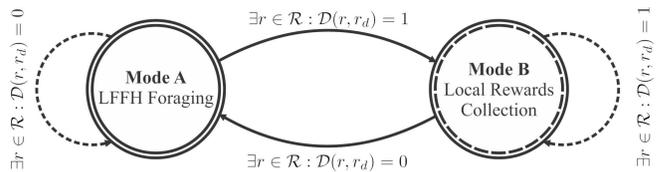}
\caption{Switching beheavior between L\'evy Flight Foraging jumps and local reward collection when targets are in detection range. }
\label{fig:hybridswitch}
\end{figure}

The outlined switching behavior is summarized through the hybrid model representation in Figure~\ref{fig:hybridswitch}. In this framework, the expression $\exists r\in \mathcal{R}: \mathcal{D}(r,r_d)=1$ indicates that there exists a reward $r$ that is within detection range $r_d$ from the agent. It is noted that while in Mode A, the foraging agent may visit the same target region multiple times, while for the purposes of the subsequently presented simulation studies the rewards may be collected only during the first visit. This assumption may be generalized to what is termed as Non-destructive Foraging (NDF)~\cite{reynolds2009adaptive,viswanathan1999optimizing}. NDF can occur either when the target sites becomes temporarily depleted or fall below some fixed concentration threshold, or when the foraged becomes satiated and leaves the area. In our results, mostly relevant to monitoring, surveillance and search of static targets, the case of Destructive Foraging (DF) is considered - the forager becomes satiated and leaves the area. In our assumption, mostly relevant to monitoring, surveillance and search of static targets, the case of Destructive Foraging (DF) is considered. During DF, the target site found by the foraging agent becomes undetectable in subsequent phases of the search (all rewards collected).

%
%

\section{SIMULATION STUDIES}\label{sec:evaluation}

In order to evaluate the applicability of the proposed approach in the framework of autonomous target search under sparse rewards considerations, a set of simulation studies were conducted. More specifically, both two-dimensional and three-dimensional simulations were considered. 

\begin{figure}[h!]
\centering
  \includegraphics[width=0.99\columnwidth]{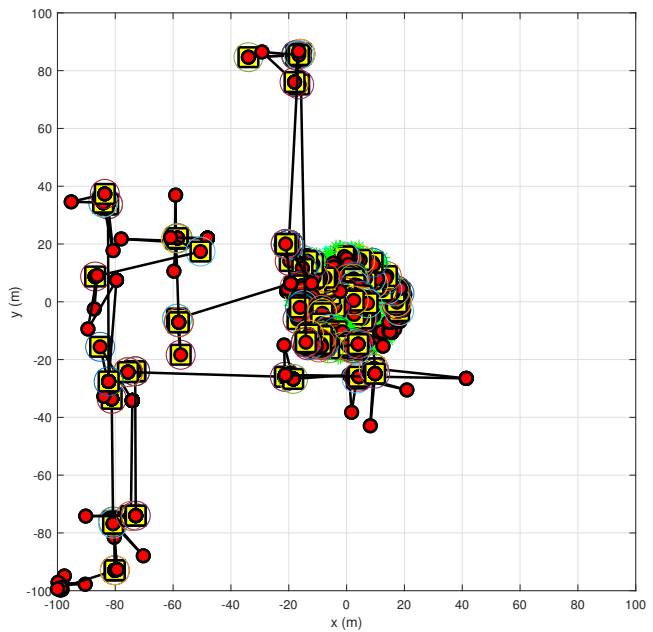}
\caption{L\'evy Flight Foraging Hypothesis-based search under sparse rewards. The robot enters from $[-100,-100]$, has a collection radius $r_d=5$ and collected the awards through random jumps respecting Equation~\ref{eq:levybasic}).  }
\label{fig:simA}
\end{figure}

\begin{figure}[h!]
\centering
  \includegraphics[width=0.99\columnwidth]{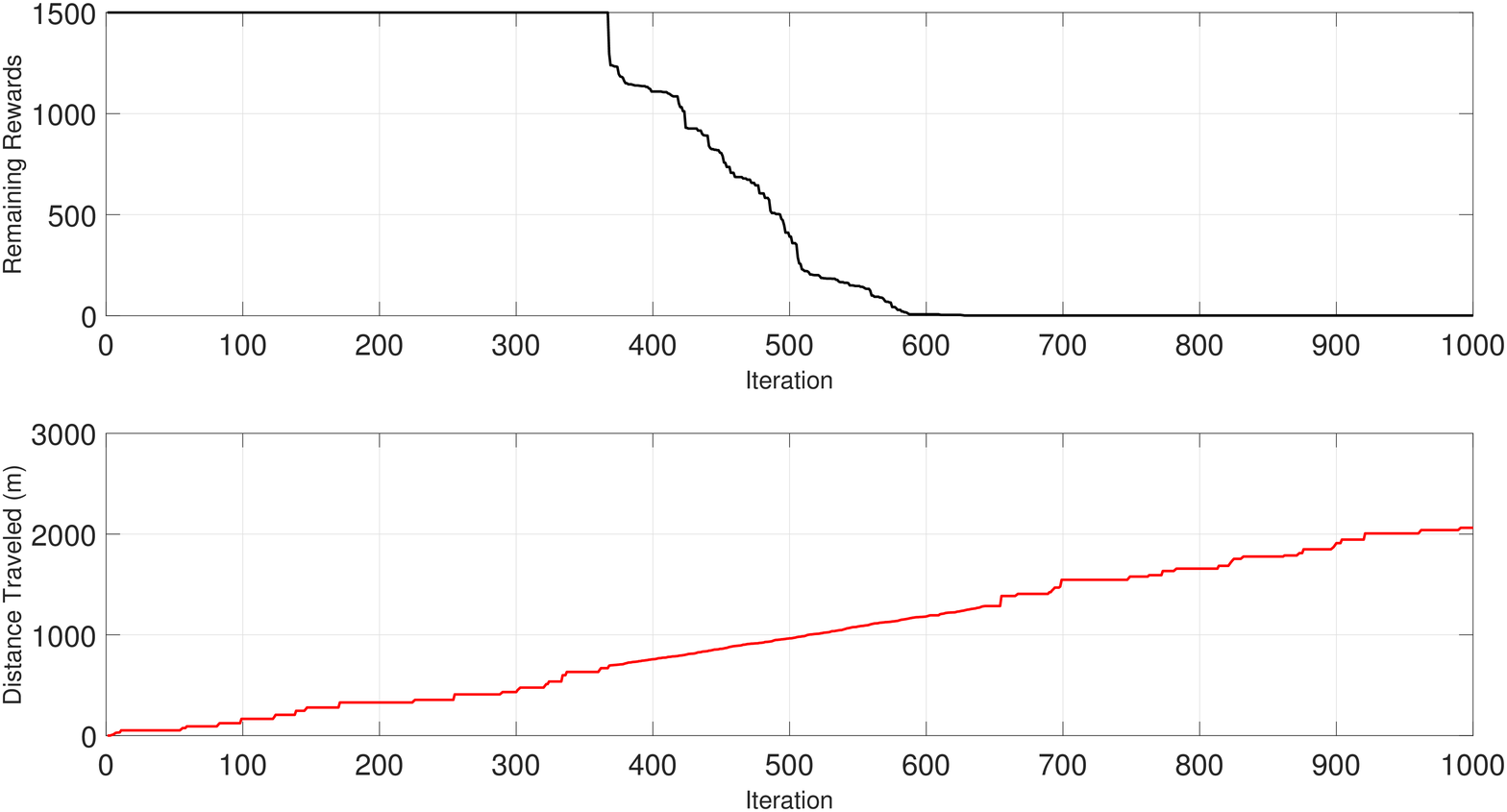}
\caption{Rewards collection rate and distance travel for L\'evy Flight Foraging Hypothesis-based search under sparse rewards. The robot enters from $[-100,-100]$, has a collection radius $r_d=5$ and collected the awards through random jumps respecting Equation~\ref{eq:levybasic}). }
\label{fig:simAanalysis}
\end{figure}

In a $2\textrm{D}$-configuration space of $100\times 100\textrm{m}$ the autonomous agent was tasked to explore and search for rewards based on target sites that were sparsely positioned. In the first case, there is a single cluster of rewards located at $[0,0]$ while the robot entered at $[-100,100]$. In total the cluster of rewards contains $1000$ rewards. An indicative run of the presented LFFH-based searcher is presented in Figure~\ref{fig:simA} alongside analysis results with respect to rewards collection rate and travelled distance in Figure~\ref{fig:simAanalysis}. 

%
%

%
%

\begin{figure}[h!]
\centering
  \includegraphics[width=0.99\columnwidth]{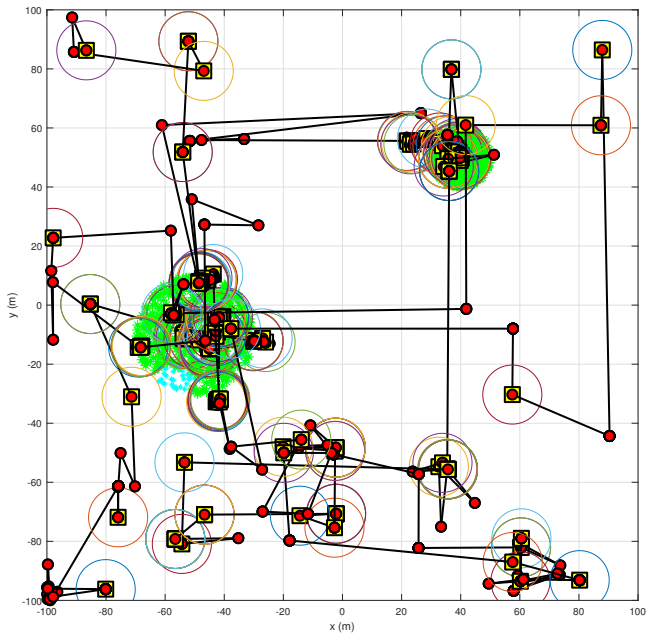}
\caption{L\'evy Flight Foraging Hypothesis-based search under sparse rewards distributed over two clusters. The robot enters from $[-100,-100]$, has a collection radius $r_d=10$ and collected the awards through random jumps respecting Equation~\ref{eq:levybasic}).  }
\label{fig:simB}
\end{figure}

\begin{figure}[h!]
\centering
  \includegraphics[width=0.99\columnwidth]{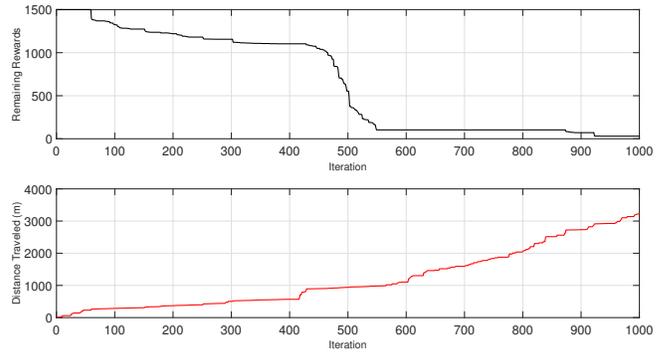}
\caption{Rewards collection rate and distance travel for L\'evy Flight Foraging Hypothesis-based search under sparse rewards distributed over two clusters. The robot enters from $[-100,-100]$, has a collection radius $r_d=10$ and collected the awards through random jumps respecting Equation~\ref{eq:levybasic}).}
\label{fig:simBanalysis}
\end{figure}

In a second $2\textrm{D}$ simulation case the autonomous agent was tasked to explore and search for rewards located in two clusters centered at $[-50,-10]$ and $[40,50]$ and rewards radius of $20$ and $10$ respectively, while in total $500,1000$ rewards are distributed at each of the clusters. The robot entered at $[-100,100]$. An indicative run of the presented LFFH-based searcher is presented in Figure~\ref{fig:simB} alongside analysis results with respect to rewards collection rate and travelled distance in Figure~\ref{fig:simBanalysis}.

%
%

%
%

\begin{figure}[h!]
\centering
  \includegraphics[width=0.99\columnwidth]{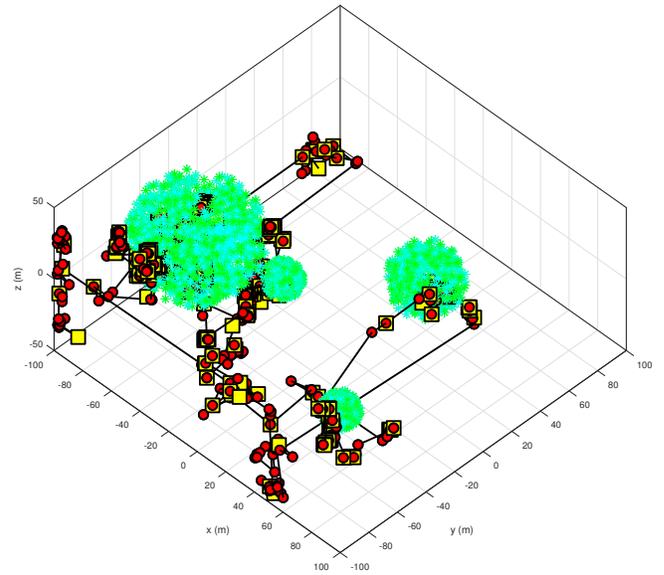}
\caption{L\'evy Flight Foraging Hypothesis-based search under sparse rewards. The robot enters from $[0,0,0]$, has a collection radius $r_d=10$ and collected the awards through random jumps respecting Equation~\ref{eq:levybasic}).  }
\label{fig:simC}
\end{figure}

\begin{figure}[h!]
\centering
  \includegraphics[width=0.99\columnwidth]{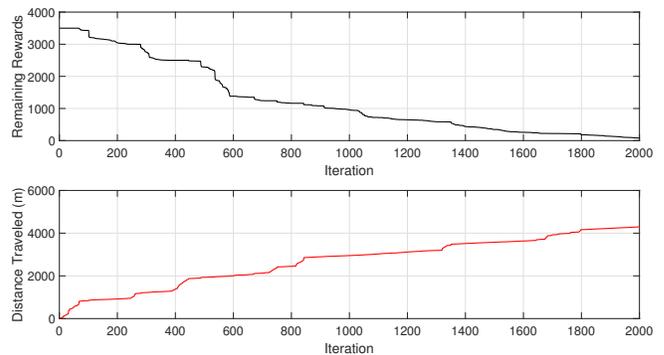}
\caption{Rewards collection rate and distance travel for L\'evy Flight Foraging Hypothesis-based search under sparse rewards distributed over two clusters. The robot enters from $[0,0,0]$, has a collection radius $r_d=10$ and collected the awards through random jumps respecting Equation~\ref{eq:levybasic}).}
\label{fig:simCanalysis}
\end{figure}

A third simulation scenario involves a $3\textrm{D}$ simulation case. The rewards are distributed over four clusters centered at $[30,30,0]$, $[-20 -20 0]$, $[-50 -50 30]$, and $[65 -65 0]$ and have rewards distribution ranges $20,10,35,10$ respectively, while in total $500, 1000, 1500, 500$ rewards are distributed per cluster. The robot entered at $[0,0,0]$. An indicative run of the presented LFFH-based searcher is presented in Figure~\ref{fig:simC} alongside analysis results with respect to rewards collection rate and travelled distance in Figure~\ref{fig:simCanalysis}.

%
%

%
%

These simulation results are representative of the performance of the proposed memoryless search method and its applicability in the case of environments with sparse rewards to be collected by the autonomous agent. 

\section{CONCLUSIONS}\label{sec:concl}

In this work, a method for cost--efficient memoryless area exploration and search under sparse information rewards is proposed. The described approach is inspired by the L\'evy Flight Foraging Hypothesis and studies confirming that a large set of species satisfy this assumption. The presented method is accompanied by a set of evaluation studies presenting its basic properties and role of usage. Future efforts will emphasize on the robotic realization of this algorithm, iterative improvements and refinements and conducting experimental results with flying robotic systems. 




\bibliographystyle{IEEEtran}
\bibliography{./BIB/LF_BIB}

\end{document}